\documentclass[letterpaper, 10 pt, conference]{ieeeconf}  

\IEEEoverridecommandlockouts                              

\usepackage{amsmath}
\usepackage{amssymb}
\usepackage{tikz}
\usetikzlibrary{positioning,calc,arrows.meta}
\usepackage{bm}
\usepackage{float}
\usepackage{booktabs}
\usepackage{multirow}
\usepackage{threeparttable}
\usepackage{color}
\usepackage{url}
\usepackage{hyperref}

\newcommand{\x}{\mathbf{x}}
\newcommand{\p}{\mathbf{p}}
\newcommand{\q}{\mathbf{q}}
\newcommand{\e}{\mathbf{e}}
\newcommand{\ome}{\boldsymbol{\omega}}
\newtheorem{definition}{Definition}
\newtheorem{remark}{Remark}
\title{\LARGE \bf
Safety-Critical Quadrotor Navigation Using Fourth-Order Control Barrier Functions with Full Nonlinear Quadrotor Dynamics
}
\author{
Yanze Liu$^\ast$, Yaşar İdikut$^\ast$, Dhruv Madhusudan Agrawal, Wei Xiao, Guanrui Li%
\thanks{$^\ast$ denotes equal contribution.}%
\thanks{Yanze Liu, Yaşar İdikut, Dhruv Madhusudan Agrawal, and Guanrui Li are with the Department of Robotics Engineering, Worcester Polytechnic Institute, Worcester, MA 01609 USA (email: \{yliu43, yidikut, dagrawal, gli7\}@wpi.edu).}%
\thanks{Wei Xiao is with Nanyang Technological University, Singapore (email: weixy@mit.edu).}%
}

\begin{document}

\maketitle
\thispagestyle{empty}
\pagestyle{empty}

\begin{abstract}
Safety-critical quadrotor navigation in cluttered environments typically relies on hierarchical architectures in which a high-level safety filter modifies position references that are then tracked by a separate attitude controller. However, this separation can create a mismatch between translational safety objectives and attitude evolution. To address this gap, we propose a safety controller based on fourth-order control barrier functions (CBFs) that explicitly incorporates the full nonlinear quadrotor dynamics, including attitude dynamics and torque-level actuation. Furthermore, we apply the integral high-order CBF technique to resolve the mixed-relative degree issue in the derived constraint formulation. 
The resulting formulation enforces obstacle-avoidance constraints while optimizing control torques, granting the safety layer direct authority over attitude behavior and eliminating the hierarchical disconnect inherent in reference-tracking safety filters. 
We validate the approaches in both high-fidelity simulation and real-world flight experiments, demonstrating reliable constraint satisfaction and successful obstacle avoidance. These results suggest that torque-level high-order CBFs provide a practical pathway to provably safer quadrotor navigation under full rigid-body dynamics.
\end{abstract}


\section*{Supplementary material}
Video:
\url{https://youtu.be/l72K2VL5B6I}
\section{Introduction}
Autonomous quadrotor flight through cluttered environments safely is a core requirement for quadrotors' application in logistics and search and rescue applications~\cite{agile,dq}. Collision-free navigation for quadrotor in such environments has been approached from several directions, including reactive avoidance \cite{reactive_filter}, optimization-based control that treats obstacles as explicit constraints \cite{ball2, cbfdrone}, and learning-based policies \cite{Zhang2024LearningVA}. Each of these layers contributes to flight safety; however, none alone can guarantee collision-free operation under all conditions the vehicle may encounter. 

On the other hand, control-level safety filters can run at the rate of the control loop and act on the vehicle dynamics directly, so they can intervene within a single control period and admit only commands the vehicle can physically execute. This makes them complementary to the approaches above. A map-based planner or learning-based planner can return a collision-free path that becomes invalid when the global map drifts, or when replanning runs slower than the vehicle moves. Under those conditions the safety filter can proactively reject a command in high frequency before it reaches the actuators, which is why we enforce the safety guarantee at the control level in this work.

Control Barrier Functions (CBFs) \cite{cbf} provide such a filter. A CBF-based quadratic program (CBF-QP) modifies the command of a nominal controller as little as possible while keeping the state inside a safe set. The standard CBF-QP applies only when the safety constraint has relative degree one \cite{control}. At higher relative degree the control input does not appear in the CBF condition, so the QP has no authority over the constraint. The $6$-degree of freedom (DoF) quadrotor dynamics fall into this case.



To address higher relative-degree constraints, High-Order Control Barrier Functions (HOCBFs) \cite{hocbf} and Exponential Control Barrier Functions (ECBFs) \cite{ecbf} address constraints of higher relative degree by constructing a sequence of constraints that ensure forward invariance of the safe set \cite{set}, thereby guaranteeing satisfaction of the original safety requirement. However, applications of this construction to quadrotors have so far stopped at the acceleration level \cite{cascade, 3rd}. The filter certifies an acceleration command, and a separate low-level controller without safety constraints is left to realize it. Because the filter never regulates the attitude dynamics, a large commanded acceleration can drive the vehicle toward an attitude the low-level controller cannot track, and aggressive maneuvers may then produce excessive tilt or even a flip. The safety certificate therefore holds for the abstraction that was filtered rather than for the quadrotor that flies.

\begin{figure}[t]
    \centering
    \includegraphics[width=\linewidth]{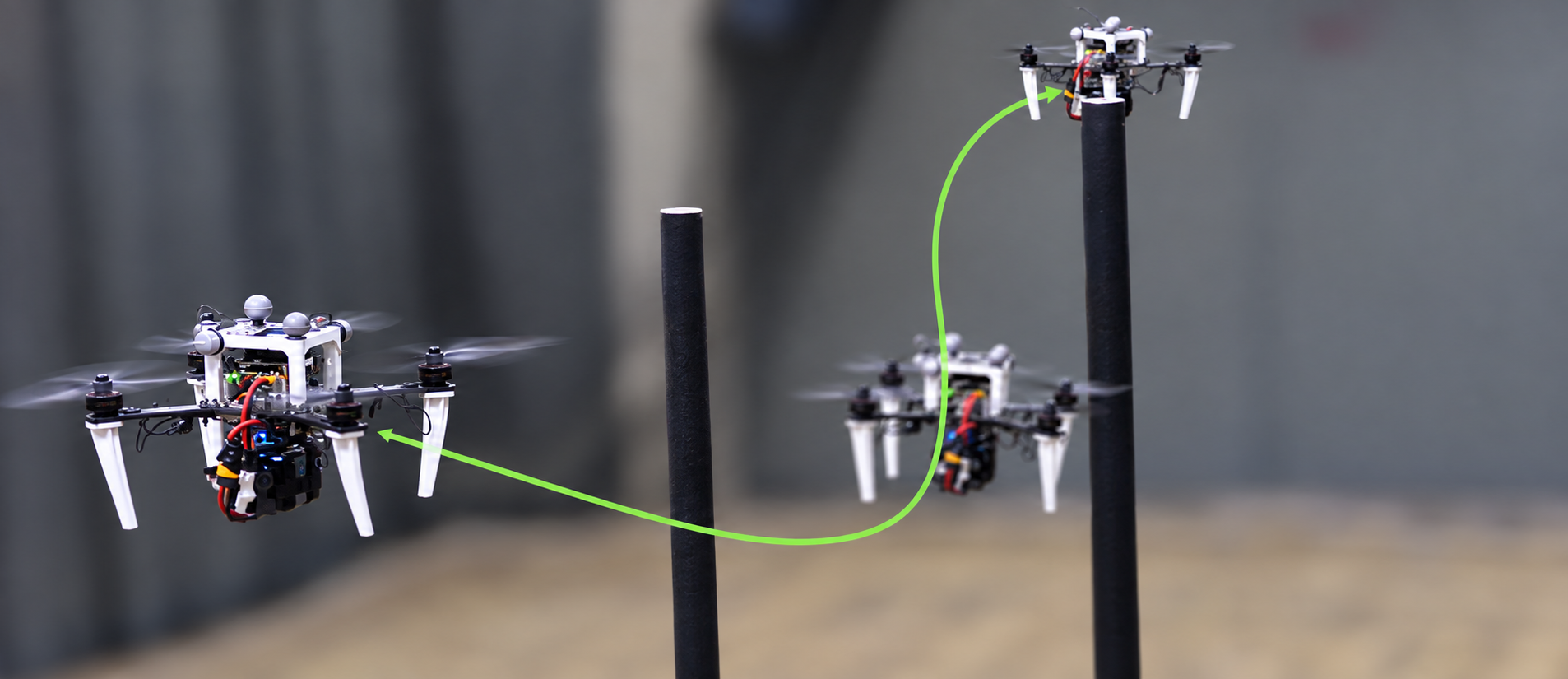}
    \caption{The quadrotor flight trajectory with the fourth-order CBF-QP for obstacle avoidance.}
    \label{fig1}
\end{figure}

To overcome this limitation, we derive a fourth-order CBF for the full nonlinear quadrotor dynamics, with the collective thrust and the three body torques as control inputs. 
We differentiate the position-dependent safety constraint until the thrust and torques appear explicitly. This lets the safety filter update thrust and torque commands and account for the attitude dynamics inside the same optimization, so a separate low-level attitude controller is no longer required. To the best of our knowledge, this is the first HOCBF formulation that enforces obstacle avoidance directly on the full nonlinear 6-DoF quadrotor dynamics at the torque level. 

The contributions of this work are as follows.
\begin{itemize}
\item We derive a fourth-order CBF condition for the full nonlinear quadrotor dynamics, in which the collective thrust and the body torques appear explicitly in the fourth derivative of a position-dependent safety constraint.
\item We applied the integral HOCBF method \cite{xiao2022multiple} to resolve the existing mixed-relative degree problem in the fourth-order CBF condition. 
\item We validate the controller in $300$ randomized cluttered environments in simulation and in hardware flight, where it records no collisions and solves within the $4\,\mathrm{ms}$ period of the onboard $250\,\mathrm{Hz}$ control loop.
\end{itemize}



\section{Related Works}
Several prior works have incorporated Control Barrier Functions (CBFs) into quadrotor control design. In \cite{3rd}, the authors employ a composite third-order CBF based on a simplified $6$-DoF rigid-body model to encode complex obstacle avoidance constraints. In that formulation, the control inputs are defined as the first-order derivatives of thrust and angular velocities, where the attitude kinematic is considered. Under this setting, an additional low-level attitude controller is still required. As a result, the attitude dynamics is not explicitly integrated into the safety filter during obstacle avoidance.
A cascaded control architecture is adopted in \cite{cascade} to ensure overall quadrotor safety. While this structure enables hierarchical safety enforcement, it increases controller complexity and introduces coupling across multiple layers. More importantly, obstacle avoidance and attitude regulation are designed independently, without an explicit analytical linkage. Such separation may lead to potential conflicts among CBF constraints when multiple safety objectives interact.
In \cite{backstep}, a CBF-based backstepping framework is proposed to impose multiple convex constraints on position, velocity, and angular rates for multirotor systems. By exploiting the strict-feedback structure, safety constraints are recursively enforced through intermediate virtual controls. However, the construction of these safe intermediate states is nontrivial and may be less transparent compared to directly formulating safety constraints on the original physical states.

\section{Preliminaries}
In this section, we introduce the preliminary background of the work proposed in this paper. First, we will present the full quadrotor dynamics in Section \ref{sec:system_dynamics}. Next, we will introduce the definition of high-order control barrier function in Section \ref{sec:hocbf}. 

\begin{figure}[t]
    \centering
    \includegraphics[width=0.8\linewidth]{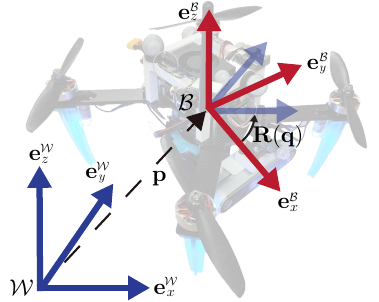}
    \caption{System convention definition: $\mathcal{W}$ and  $\mathcal{B}$ denote the inertial frame and the body frame respectively.}
    \label{fig:system_representation}
\end{figure}

\subsection{System Dynamics}\label{sec:system_dynamics}
We first define the inertial and world frames as $\mathcal B = \{\e_x^{\mathcal{B}}, \e_y^{\mathcal{B}},\e_z^{\mathcal{B}}\}$ and $\mathcal W = \{\e_x^{\mathcal{W}}, \e_y^{\mathcal{W}},\e_z^{\mathcal{W}}\}$ as in Fig. \ref{fig:system_representation}, respectively. 
Then, consider the standard quadrotor model, where the quadrotor states are defined as $\mathbf{x} = \begin{bmatrix} \mathbf{p}^\top & \mathbf{v}^\top & \mathbf{q}^\top & \boldsymbol{\omega}^\top \end{bmatrix}^\top \in \mathbb{R}^{13}$. $\mathbf{p} = [p_x, p_y, p_z]^\top \in \mathbb{R}^3$ denotes the quadrotor position in the inertial frame, $\mathbf{v} = [v_x, v_y, v_z]^\top \in \mathbb{R}^3$ is the velocity, $\mathbf{q} = [q_w, q_x, q_y, q_z]^\top \in \mathbb{S}^3$ represents the unit quaternion denoting the quadrotor's attitude of the drone in the inertial frame, and $\boldsymbol{\omega} = [\omega_{x}, \omega_{y}, \omega_{z}]^\top \in \mathbb{R}^3$ is the angular velocity in the inertial frame. The quadrotor dynamics is shown as follows:
\begin{subequations}\label{eq:dynamics}
\begin{align}
    &\dot {\p} = \mathbf v ,\,\,\, \dot {\mathbf v} = \frac{f_b}{m} \mathbf{R}(\q)\mathbf{e}_z^{\mathcal W} - g\mathbf{e}_z^{\mathcal W},\\
    &\dot \q = \frac{1}{2}\mathbf\Omega(\ome)\q, \,\,\,\dot \ome = \mathbf I^{-1}\left(\mathbf\tau-\ome\times (\mathbf I\ome)\right),\label{dotwb}
\end{align}
\end{subequations}
where the control inputs $\mathbf{u} = [f_b, \tau]^\top \in \mathbb{R}^4$ contains the thrust $f_b$ and the torque $\tau = [\tau_x, \tau_y,\tau_z]^\top$, 
the gravitational acceleration is $g = 9.81~\text{m/s}^2$, $m$ is the quadroter mass, $\mathbf{I} = \text{diag}(I_{xx}, I_{yy}, I_{zz})$ is the inertia matrix, the vector $\mathbf{e}_z^{\mathcal W} = [0, 0, 1]^\top$, and the cross product is denoted as $\times$ in \eqref{dotwb}. The rotation matrix $\mathbf R$ from the body frame to the inertial frame is:
\begin{equation*}
\mathbf{R}(\mathbf{q}) = \scalebox{0.8}{$\begin{bmatrix}
1-2(q_y^2+q_z^2) & 2(q_xq_y - q_zq_w) & 2(q_xq_z + q_yq_w) \\
2(q_xq_y + q_zq_w) & 1-2(q_x^2+q_z^2) & 2(q_yq_z - q_xq_w) \\
2(q_xq_z - q_yq_w) & 2(q_yq_z + q_xq_w) & 1-2(q_x^2+q_y^2)
\end{bmatrix}$}.
\end{equation*}
The skew-symmetric matrix of the quadrotor angular velocity $\boldsymbol{\omega}$, $\boldsymbol{\Omega}(\boldsymbol{\omega})$, is shown as follows
\begin{equation*}
\boldsymbol{\Omega}(\boldsymbol{\omega}) = \begin{bmatrix}
0 & -\omega_{x} & -\omega_{y} & -\omega_{z} \\
\omega_{x} & 0 & \omega_{z} & -\omega_{y} \\
\omega_{y} & -\omega_{z} & 0 & \omega_{x} \\
\omega_{z} & \omega_{y} & -\omega_{x} & 0
\end{bmatrix}.
\end{equation*}
Note that the system dynamics can be expressed in control-affine form: $\dot{\mathbf{x}} = \mathbf{f}(\mathbf{x}) + \mathbf{g}(\mathbf{x})\mathbf{u}$, where the drift vector field $\mathbf{f}(\mathbf{x})$ and control matrix $\mathbf{g}(\mathbf{x})$ can be derived from \eqref{eq:dynamics}.


\subsection{High-Order Control Barrier Function}\label{sec:hocbf}
Consider the control affine system mentioned above with the control input constraint $\mathbf u\in\mathcal U$. Several definitions of key components related to the high-order control barrier functions are given below. 
\begin{definition}[Forward Invariance]
A set $\mathcal C$ is forward invariant for the system \eqref{eq:dynamics}
if $\x(t)\in\mathcal C$ for all $t\ge 0$ whenever $\x(0)\in\mathcal C$  \cite[Definition 4.1]{set}. \label{def:forward_invariance}
\end{definition}
Based on Definition \ref{def:forward_invariance}, we define a barrier function $h(\mathbf{x})$ for obstacle avoidance such that the corresponding safe set is:
\begin{equation}
\mathcal{C} = \{\x \in \mathbb{R}^{13} : h(\x) \geq 0\}.
\label{eq:safe_set}
\end{equation}
\begin{definition}[Control Barrier Function \cite{cbf}]
    Given a set $\mathcal{C}$ as in \eqref{eq:safe_set}, a continuously differentiable $h(\x)$ is a candidate CBF for the system in \eqref{eq:dynamics} if there exists an extended class $\mathcal K$ function $\alpha$ s.t.
    \begin{align}
        \sup_{u\in\mathcal U} [L_{\mathbf f}h(\x)+L_{\mathbf g}h(\x)\mathbf u+\alpha(h(\x))]\geq 0,\quad \x\in\mathcal{C},
    \end{align}
    where $L_{\mathbf f}$ and $L_{\mathbf g}$ denote the Lie derivative along $\mathbf f(\x)$ and $\mathbf g(\x)$ defined previously. 
\end{definition}

To enforce the CBF condition for the system defined in \eqref{eq:dynamics}, it is essential to examine the relative degree of the safety constraint with respect to the system dynamics.
In particular, the maximum relative degree of a safety constraint $h(\x)$ with respect to the system \eqref{eq:dynamics} is $4$. This arises when the constraint depends only on the position, i.e., $h(\p)$, since four time derivatives are required before the control torque $\tau$ explicitly appears in the resulting expression. We consider this case for \eqref{eq:dynamics} in the following. 
Define a sequence of scalar functions $h_i : \mathbb R^{13} \to \mathbb{R}$ for $i \in \{1, \ldots, 4\}$ recursively as 
\begin{align}\label{hs}
h_i(\x) = \dot{h}_{i-1}(\x) + \alpha_i\big(h_{i-1}(\x)\big), 
\quad i \in \{1, \ldots, 4\},
\end{align}
with the initialization $h_0(\x) := h(\x)$ defined as above. 
Here, each $\alpha_i(\cdot)$, $i \in \{1, \ldots, 4\}$, is an extended class $\mathcal{K}$ function that is $(4-i)$ times continuously differentiable.

Based on this construction, defining the corresponding safe sets $C_i$, $i \in \{1, \ldots, m\}$, associated with $h_i$, defined by
\begin{align}\label{sets}
C_i := \left\{ \x\;:\; h_{i-1}(\x) \geq 0 \right\}, 
\quad i \in \{1, \ldots, 4\}.
\end{align}
\begin{definition}[High Order Control Barrier Function \cite{hocbf}]
Let the sets $C_i$, $i \in \{1, \ldots, 4\}$, be given by \eqref{sets}, and let the functions $h_i(\x)$, $i \in \{1, \ldots, 4\}$, be defined as in \eqref{hs}. 
The function $h:\mathbb{R}^{13} \to \mathbb{R}$ is called a High Order Control Barrier Function (HOCBF) of relative degree $4$ for system \eqref{eq:dynamics} with respect to the torque $\tau$ if there exist extended class $\mathcal{K}$ functions $\alpha_i$, $i \in \{1, \ldots, 3\}$, each $(4-i)$ times continuously differentiable, together with an extended class $\mathcal K$ function $\alpha_4$, such that
\[
\sup_{\boldsymbol{u} \in \mathcal U} 
\left(
L_{\mathbf f} h_{3}(\x) 
+ L_{\mathbf g} h_{3}(\x) \boldsymbol{u} 
+ \alpha_4\big(h_{3}(\x)\big)
\right) 
\ge 0,
\]
for every $\x \in C_1 \cap \cdots \cap C_4$. 
In addition, it is required that $L_{\mathbf g} h_{3}(\x) \neq 0$ whenever $h(\x) = 0$.
\end{definition}

\section{HOCBF with Full Quadrotor Dynamics}\label{sec:4ocbf}
In this section, we present the mathematical derivation of the high-order control barrier function that leverages the full quadrotor dynamics model introduced in \eqref{eq:dynamics}. 
By differentiating the control barrier function $h$ four times, the control input $\mathbf{u}$ that includes the thrust and torque of the quadrotor appears completely in the HOCBF constraints. The resulting HOCBF constraints, which are affine in $\mathbf{u}$, are embedded in a convex QP to guarantee safety. 

\subsection{High-order CBF Derivation}
Consider a safety set that encodes the ball obstacle avoidance constraint: $\mathcal{C} = \{\x|\|\mathbf e_p\|^2-r^2\geq0\}$ given the corresponding CBF 
\begin{align}\label{cbf}
h(\x) = \|\mathbf e_p\|^2-r^2,
\end{align} 
where the position error is denoted as $\mathbf e_p = \p-\mathbf c$, and the ball obstacle is defined by the center $\mathbf c = [c_x,c_y,c_z]^\top$ and the radius $r$. Note that such a design is widely applicable in multiple scenarios \cite{ball1,ball2}. 
Each order of the CBF is derived in the following according to \eqref{hs}. The first-order CBF is derived as 
\begin{align}
    h_1(\x) = 2\mathbf{e}_p^\top\mathbf{v} + k_1h,
\end{align}
and the second-order CBF is derived as
\begin{align}\label{2nd}
    h_2(\x, f_b) = 2\|\mathbf{v}\|^2 + 2\mathbf{e}_p^\top\mathbf{a}
             + 2k_1\mathbf{e}_p^\top\mathbf{v}+ k_2h_1,
\end{align}
where the acceleration is $\mathbf{a} = \frac{f_b}{m}\mathbf{R}(\q)\mathbf{e}_z^{\mathcal W} - g\mathbf{e}_z^{\mathcal W} $. Note that the first control input, the total thrust $f_b$, appears in this order. The third-order CBF is derived as
\begin{align}
      h_3(\x, f_b) \!=& 6\mathbf{v}^\top\!\mathbf{a}
             \hspace{-0.3em}+\hspace{-0.3em} 2\mathbf{e}_p^\top\mathbf{j}
             \!+\hspace{-0.3em} 2\hspace{-0.1em}(k_1\hspace{-0.4em}+\hspace{-0.3em}k_2)
               \bigl(\|\mathbf{v}\|^2
                     \!+\! \mathbf{e}_p^\top\mathbf{a}\bigr)\nonumber\\
         &\hspace{-0.3em}+\hspace{-0.3em} 2k_1k_2\mathbf{e}_p^\top\mathbf{v}
             \hspace{-0.3em}+\hspace{-0.3em} k_3h_2,
\end{align}
where the jerk is defined as the derivative of the acceleration 
\begin{equation}
\mathbf{j} = \dot{\mathbf{a}} = \frac{\bar {f_b}}{m}\mathbf{R}\mathbf{e}_z^{\mathcal W} + \frac{f_b}{m}\mathbf{R}(\boldsymbol{\omega} \times \mathbf{e}_z^{\mathcal W}),
\end{equation}
where $\bar {f_b} = \dot{f_b}$. 
Finally, we derive the fourth-order CBF as follows:
\begin{align}\label{4cbf}
  h_4(\x,\mathbf u) &= 6\|\mathbf{a}\|^2
             + 8\mathbf{v}^\top\mathbf{j}
             + 2\mathbf{e}_p^\top\mathbf{s} \nonumber\\
         &\quad
             + 6(k_1+k_2+k_3)\mathbf{v}^\top\mathbf{a}
             + 2(k_1+k_2+k_3)\mathbf{e}_p^\top\mathbf{j}
             \nonumber\\
         &\quad
             + 2(k_1k_2+k_1k_3+k_2k_3)
               \bigl(\|\mathbf{v}\|^2+\mathbf{e}_p^\top\mathbf{a}\bigr)
             \nonumber\\
         &\quad
             + 2k_1k_2k_3\,\mathbf{e}_p^\top\mathbf{v}
             + k_4h_3,
\end{align}
where the snap is as follows, given the definition of $\dot\ome$ in \eqref{dotwb}: 
\begin{align}
  \mathbf{s} &= \ddot{\mathbf{a}} =\frac{\tilde{f_b}}{m}\mathbf{R}\mathbf{e}_z+\frac{2\bar{f_b}}{m}\mathbf R(\ome\times \e_z)\nonumber\\
  &+ \frac{f_b}{m}\mathbf{R}\Bigl[
    \boldsymbol{\omega}{\times}(\boldsymbol{\omega}{\times}\mathbf{e}_z^{\mathcal W})
    + \bigl(\mathbf{I}^{-1}({\boldsymbol{\tau}}
      - \boldsymbol{\omega}{\times}\mathbf{I}\boldsymbol{\omega})\bigr)
      {\times}\,\mathbf{e}_z^{\mathcal W}\Bigr],
\end{align}
where $\tilde {f_b} = \ddot{f_b}$. 

Note that all four variables of control input $\mathbf{u} = [f_b, \tau_{x}, \tau_{y}, \tau_{z}]$ appears 
in $h_4$, confirming that the fourth-order condition is the lowest-order 
constraint as an affine function of all the control inputs given the CBF form \eqref{cbf}. 

\subsection{Integral HOCBF Design}

Although all control inputs appear explicitly in the fourth-order barrier condition \eqref{4cbf}, it is still a problem how to handle the derivative of the control input $\bar{f_b}$ and $\tilde {f_b}$. To resolve this issue, we adopt the integral-HOCBF formulation proposed in \cite{ames2020integral, xiao2022multiple}. We first determine the relative degree set as $S = \{2, 4\}$ of the function $h(\x)$ for the system \eqref{eq:dynamics}, where the relative degree w.r.t. the thrust $f_b$ is $2$, and the torque is $4$. 
Then, we define the auxiliary state as $\mathbf {f_b} = [f_b, \bar {f_b}]$, the augmented system becomes
\begin{subequations}\label{eq:aug_dynamics}
\begin{align}
    &\dot {\p} = \mathbf v \\
    &\dot {\mathbf v} = \frac{f_b}{m} \mathbf{R}(\q)\mathbf{e}_z^{\mathcal W} - g\mathbf{e}_z^{\mathcal W}\\
    &\dot \q = \frac{1}{2}\mathbf\Omega(\ome)\q\\
    &\dot \ome = \mathbf I^{-1}\left(\mathbf\tau-\ome\times (\mathbf I\ome)\right),\\
    &\dot f_b = \bar {f_b},\\
    &\dot {\bar {f_b}} = \tilde{f_b}, 
\end{align}
\end{subequations}
given the augmented state $[\p, \mathbf v,\q,\ome,\mathbf {f_b}]$ and the control input $[\tilde {f_b}, \mathbf \tau]$. We can set the initial augmented state as $[f_b(0), \bar{f_b}(0)] = [mg, 0]$. 
Then, the final HOCBF condition can be expressed as
\begin{align}\label{cbfconstraint}
    L_\mathbf{f}^4h+L_{\mathbf{g}_{\tilde f_b}}L_{\mathbf f}^3h\tilde f_b +\sum_{i=x,y,z} L_{\mathbf g_{\tau_i}}L_{\mathbf f}^3h\tau_i+\alpha(\eta)\geq 0,
\end{align}
where $\eta = [h,h_1, h_2, h_3]^\top$; $\mathbf{g}_{\tilde f_b}$ and $\mathbf g_\tau$ are the corresponding system part of $\tilde f_b$ and $\tau$ in \eqref{eq:aug_dynamics}, respectively. 
\begin{figure}
    \centering
    \includegraphics[width=\linewidth]{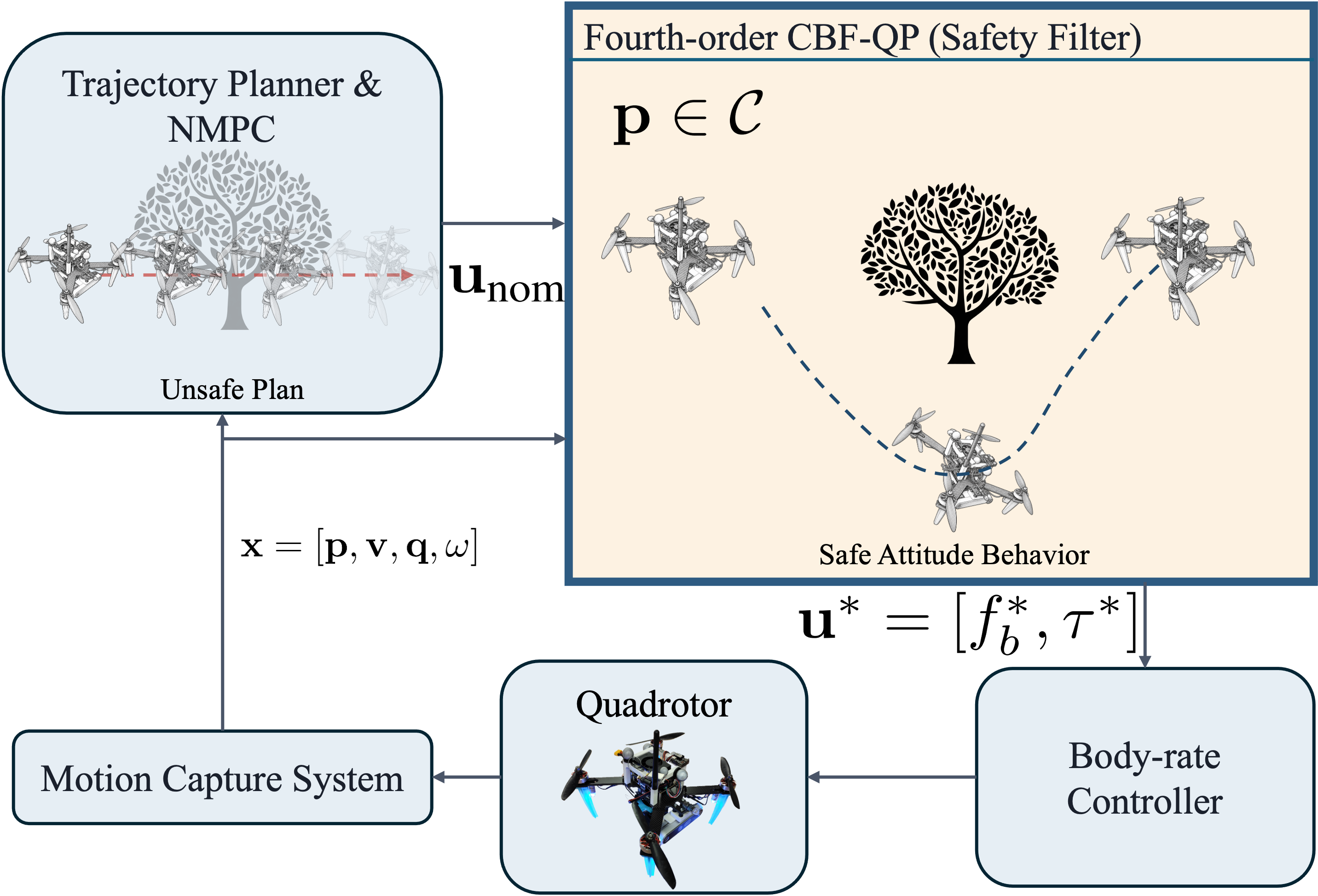}
    \caption{The cascaded control structure: the nominal NMPC controller and the CBF-QP safety filter}
    \label{fig:control_structure}
\end{figure}
\section{Control Design with HOCBF}\label{sec:nom}
In this section, we are going to implement the derived fourth-order CBF \eqref{cbfconstraint} with the safety filter for a nominal NMPC controller, and we call it as the cascaded structure.   
The safety filter solves the following convex QP at each timestep:
\begin{subequations}\label{eq:cbf_qp}
\begin{align}
    \min_{\mathbf u\in \mathcal{U}} \quad &
    \left\| \mathbf u - \mathbf u_{\text{ref}} \right\|^2 \\
    \text{s.t.} \quad
    L_\mathbf{f}^4h+L_{\mathbf{g}_{\tilde f_b}}L_{\mathbf f}^3h&\tilde f_b +\sum_{i=x,y,z} L_{\mathbf g_{\tau_i}}L_{\mathbf f}^3h\tau+\alpha(\eta)\geq 0,
\end{align}
\end{subequations}
where $\mathbf u_{\text{ref}}$ is the nominal control input supplied by the NMPC shown later. 
After deriving $\bar f_b$ and $\tilde f_b$, we can do the integral to recover the value of the original thrust 
\begin{align}\label{int_ctrl}
    f_b(t) = f_b(0) +\bar{f_b}(0)t+\int_0^t (t-\tau)\tilde {f_b}(\tau)d\tau
\end{align}
In practice, we can solve \eqref{int_ctrl} by having the discretization form for $k\in\mathbb Z_+$: 
\begin{subequations}\label{prac_ctrl}
\begin{align}
    &f_{b,k+1} = f_{b,k} +T\bar{f}_{b,k} + \frac{1}{2}T^2\tilde {f}_{b,k}\\
    &\bar f_{b, k+1} = \bar f_{b,k} + T\tilde f_{b,k},
\end{align}
\end{subequations}
where $T>0$ denotes the sampling time.  
Note that the filter is agnostic to the choice of nominal controller; readers can also use other controllers to provide the reference control inputs. The proposed QP minimizes deviation from the planned input while guaranteeing that the applied control keeps the system within the safe set $\mathcal{C}$.

\begin{remark}
The safety controller \eqref{eq:cbf_qp} is designed for the continuous-time system, while the modern robotics design is based on the discrete-time signal. However, the disturbance induced by this gap is negligible, as shown in the later experiment section. Related works can be found in \cite{sampled1} \cite{sampled2}. 
\end{remark}

At each time step, the NMPC solves a finite-horizon optimal control problem over a prediction horizon of $N$ steps with the sampling period $T$ at $k\in\mathbb Z_{\geq0}$:
\begin{subequations}\label{eq:mpc}
\begin{align}
    \min_{\{\mathbf u_k\}_{k=0}^{N-1}}  &
    \sum_{k=0}^{N-1} 
    \left(
        \left\|\x_k - \x_k^{\text{ref}}\right\|_Q^2+\left\|\mathbf u_k - \mathbf u_k^{\text{ref}}\right\|_R^2\right)
    + \\&~~~~~~~~~~~~~~~~~~~~~~~~~~~~~~~~
    \left\|\x_N - \x_N^{\text{ref}}\right\|_P^2 \nonumber\\
\text{s.t.} \quad 
& \x_{k+1} = f_{d}(\x_k, \mathbf u_k), \label{dsys}\\
& \x_k \in \mathcal{X}, \quad \mathbf u_k \in \mathcal{U},
\end{align}
\end{subequations}
where $Q\in \mathbb S_+^{13}$, $R\in \mathbb S_+^{4}$, and $P\in \mathbb S_+^{13}$ are the positive definite matrices penalizing deviations from the reference trajectory in position, velocity, attitude, and angular rates; the reference state $\x^{\text{ref}}$ and input $\mathbf u^{\text{ref}}$ are given by a trajectory planner based on the differential flatness properties of a quadrotor. The state and input constraints $\mathcal{X}$ and $\mathcal{U}$ are assumed to be box constraint which encodes some inherent requirements of the quadrotor system.  
Finally, the system $f_d$ here denotes the discretized form of \eqref{eq:dynamics} using a zero-order hold (ZOH) assumption for $k\in\mathbb Z_{\geq 0}$ expressed as
\begin{align}
    \x_{k+1} = \x_{k} + \int_{kT}^{(k+1)T}\big(\mathbf f(\x_{t}) + \mathbf g(\x_{t})\mathbf u_{k}\big)dt,
\end{align}
where \(\x_{k} = \x(kT)\). The control input \(\mathbf u_{k}\) is held constant over each sampling interval according to the ZOH condition, that is, \(\mathbf u(t) = \mathbf u(kT)\) for \(t \in [kT, (k+1)T)\). We can assume that the Lipschitz continuity of \(\x(t)\) is preserved under this discretization.

\section{Experimental Results}
We evaluate the proposed fourth-order CBF-QP on quadrotor obstacle avoidance tasks in both simulation and real-world environments. Section \ref{sec:setup} describes the controller configuration shared by all experiments. Section \ref{sec:sim} reports $300$ randomized simulation trials, and Section \ref{sec:real-world} reports the hardware flight.


\begin{figure*}[t]
    \centering
    \includegraphics[width=\textwidth]{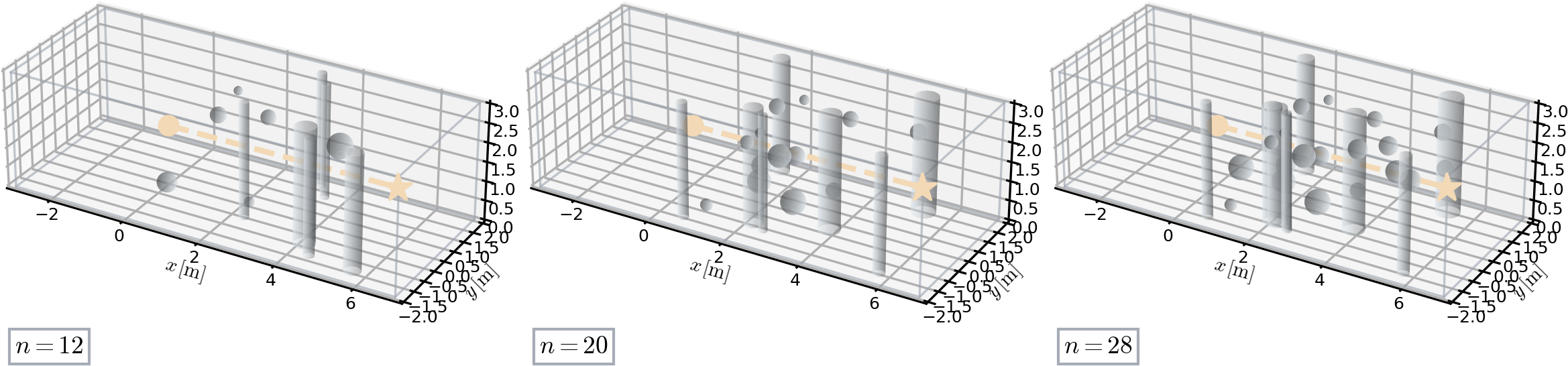}
    \caption{One example obstacle layout at each clutter level, with $n=12$ (top), $n=20$ (middle), and $n=28$ (bottom). The red dashed line is the minimum-snap reference trajectory, the red disk marks the start position, and the red star marks the goal. Green cylinders and orange spheres are obstacles, and the gray box is the flight volume enforced by the boundary constraints.}
    \label{fig:mujocoenv}
\end{figure*}

\begin{figure}[t]
    \centering
    \includegraphics[width=1\linewidth]{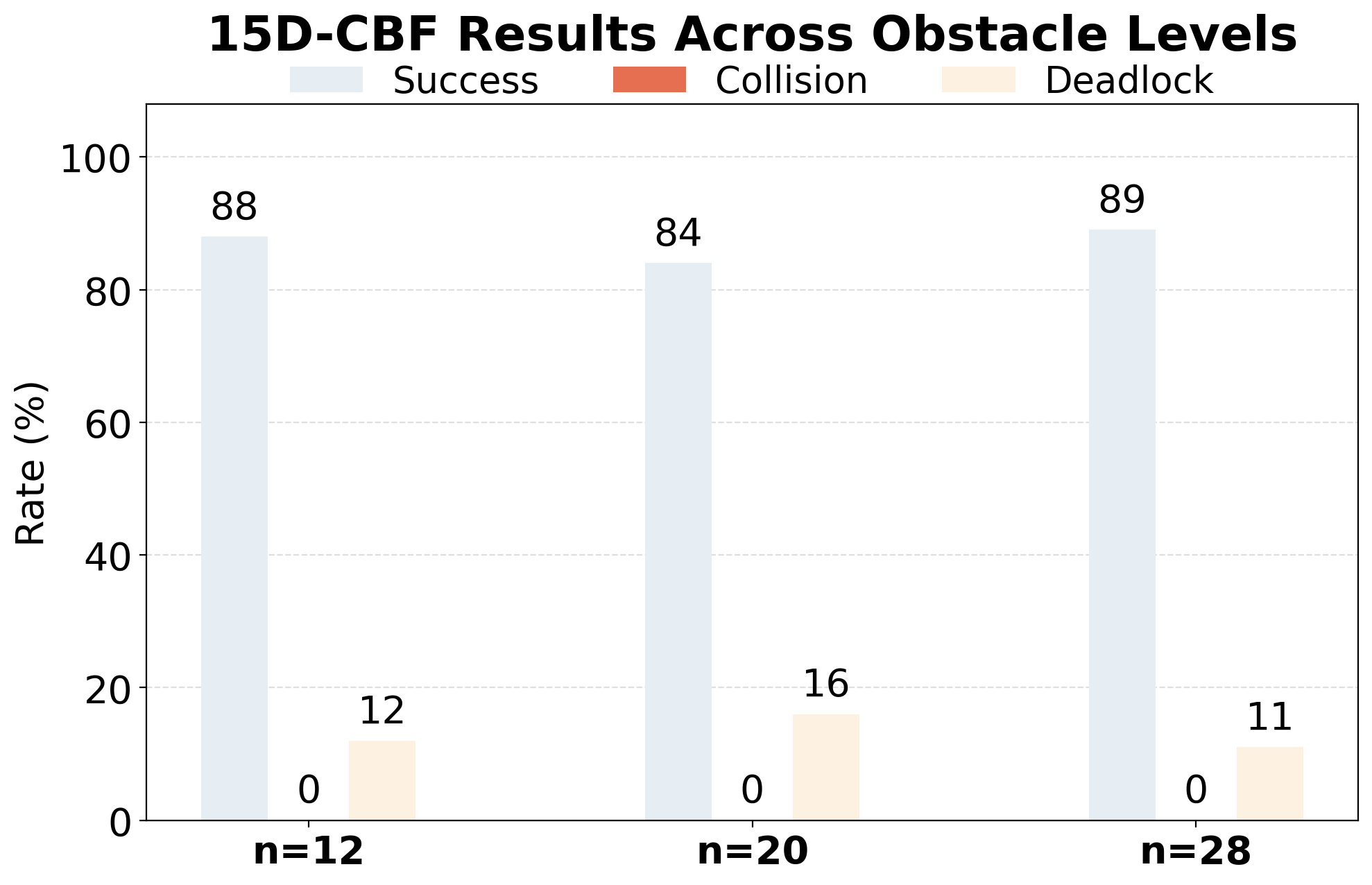}
    \caption{\textbf{The fourth-order CBF-QP records no collision at any clutter level.} Outcome of $100$ randomly generated environments per obstacle count $n$, reported as the percentage of trials ending in success, collision, and deadlock. A trial counts as a success when the quadrotor reaches the goal region without contacting an obstacle. Every remaining trial is a deadlock, in which the quadrotor stops on an obstacle boundary without violating the safety constraint.} 
    \label{fig:statistical_result}
\end{figure}

\subsection{Experiment Setups} \label{sec:setup}
All experiments use the cascaded architecture shown in Fig. \ref{fig:control_structure}, in which the NMPC generates the nominal command and the fourth-order CBF-QP filters it before actuation. The NMPC uses a prediction horizon of $N=15$ and a sampling period $T = 0.1\,\mathrm s$. 

Every obstacle $j$ contributes one safety constraint of the form given in \eqref{cbf}, $h(\x) = \|\mathbf e_p^j\|^2-(r^j)^2$ for $j=1,\cdots, J$, where $J$ is the number of total obstacles. The NMPC model \eqref{dsys} is subject to the state and input constraints
\begin{align*}
&\mathcal X = \{ \q|q_w,q_x,q_y,q_z\in[-1,1]\}\\
&\mathcal{U} = \{(f_b,\tau)|f_b\in[0, 39], \tau_x,\tau_y,\tau_z\in[-0.1, 0.1]\},
\end{align*}
where the thrust $f_b$ is in $\mathrm N$ and the torque  $\tau_x,\tau_y,\tau_z$ is in $\mathrm N\,\mathrm m$. 

We solve the nominal control \eqref{eq:mpc} and the safety filter with \texttt{acados} \cite{acados} using the sequential quadratic programming real-time iteration scheme \cite{rti}. The NMPC tracks the reference trajectory but does not enforce the obstacle-avoidance constraint $\mathcal{C}$, so the proposed fourth-order CBF-QP \eqref{eq:cbf_qp} filters the nominal command before it reaches the vehicle's low-level autopilot. 

\begin{figure}[t]
    \centering
    \includegraphics[width=1\linewidth]{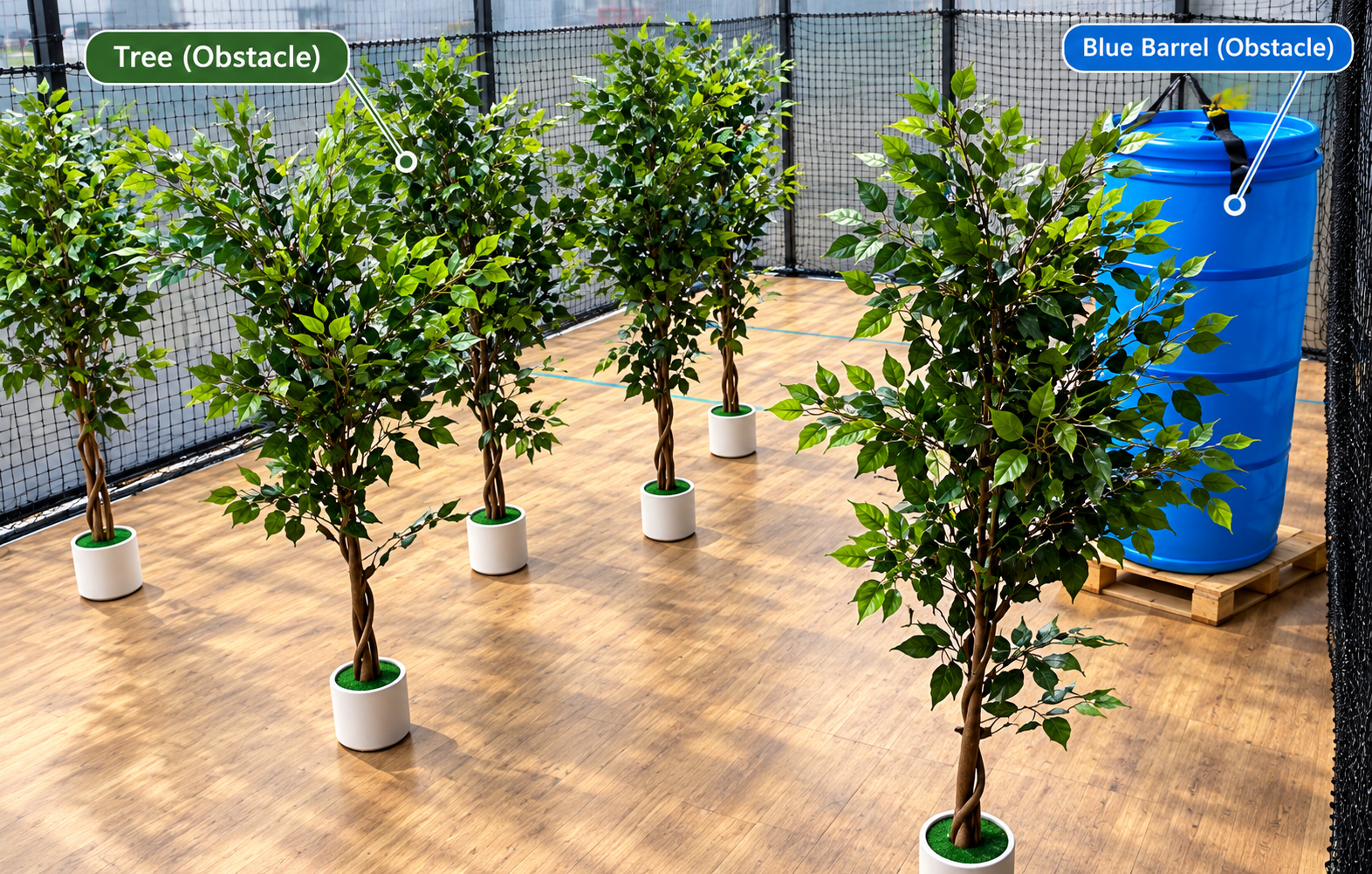}
    \caption{Schematic illustration of the cluttered real-world experiment environment.}
    \label{obs_env}
\end{figure}
\subsection{Simulation Results}\label{sec:sim}

\subsubsection{Environment Setup}
We evaluate the controller in $300$ randomly generated three-dimensional environments simulated in MuJoCo~\cite{mujoco}. The environments span three clutter levels defined by the obstacle count $n$, with $n=12$, $n=20$, and $n=28$, and each level contains $100$ environments. Every environment mixes spherical and cylindrical obstacles, and Fig. \ref{fig:mujocoenv} shows one example layout at each level. In each trial the quadrotor flies from $\mathbf{p}_0 = (0, 0, 1.5)^\top\,\mathrm m$ to the goal $\mathbf{p}_f = (6, 0, 1.5)^\top\,\mathrm m$ along a minimum-snap reference trajectory \cite{snap}.



We generate each environment from a fixed random seed, and the shape of every obstacle is drawn from the spherical and cylindrical types with equal probability. 
Obstacle radii are sampled uniformly from $0.1$ to $0.3\ \mathrm m$, and obstacle centers are sampled within a $5\ \mathrm m$ diameter tube around the nominal start-to-goal trajectory. 
We reject and resample any layout that places an obstacle too close to the goal, blocks every feasible path after inflation by the safety margin, or leaves the straight-line trajectory unobstructed. 
This rejection step yields reproducible environments in which obstacle avoidance is both necessary and possible.
\begin{figure*}[t]
    \centering
    \includegraphics[width=1\linewidth]{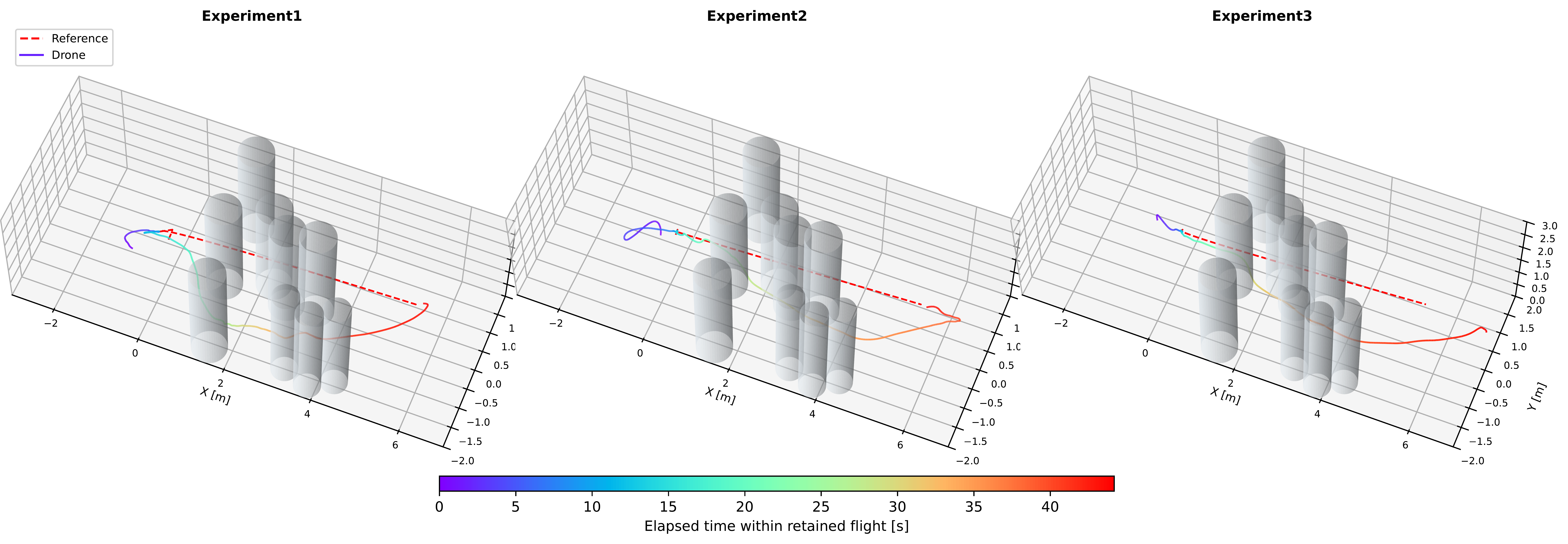}
    \caption{Three-dimensional visualization of the three measured quadrotor trajectories and the cluttered obstacle environment.}
    \label{fig:3d}
\end{figure*}

\begin{figure*}[t]
    \centering
    \includegraphics[width=1\linewidth]{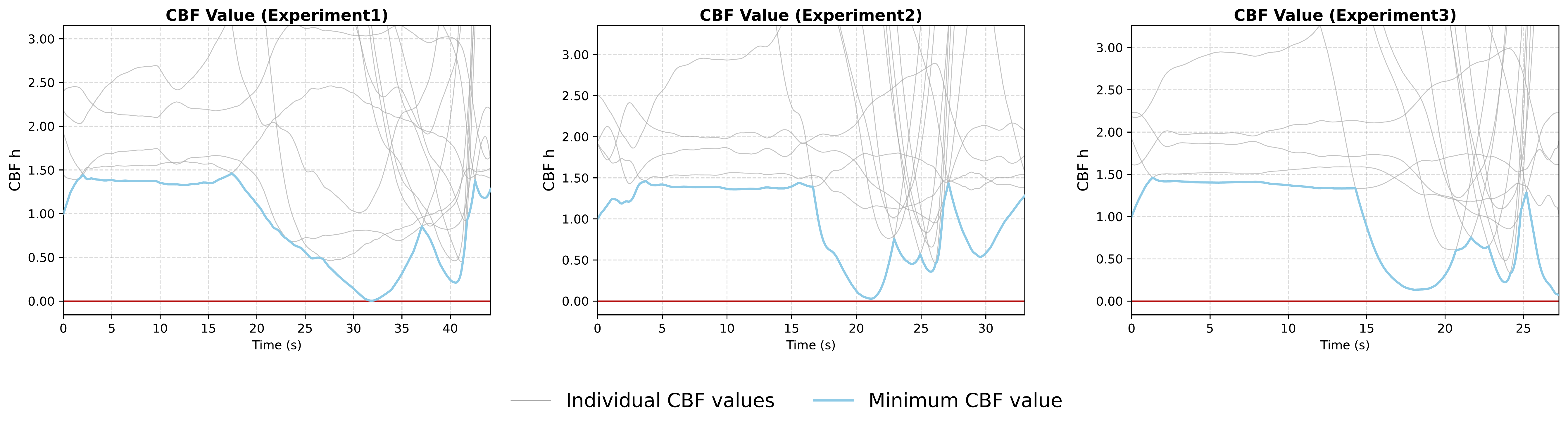}
    \caption{Comparison of the CBFs value $h$ across three flights.}
    \label{fig:cbf_cons}
\end{figure*}
In addition, six axis-aligned half-space constraints confine the quadrotor to the flight volume $p_x=[-3,\, 7]\ \mathrm m$, $p_y=[-2,\, 2]\ \mathrm m$, $p_z=[0,\, 3]\ \mathrm m$ throughout each trial. 
The filter therefore enforces $n$ obstacle constraints and six boundary plane constraints in every trial.



\subsubsection{Parameter Selection}
We take every extended class $\mathcal K$ function to be linear, so that $\alpha_i(h_{i-1}(\x)) = \kappa_i h_{i-1}(\x)$ for $i=1,\cdots, 4$. The gains are tuned per obstacle type and are listed from the first to the fourth order. We use $\kappa_{\text{cylinder}}=[1.8, 5.8, 10.0, 7.0]$ for cylinders, $\kappa_{\text{sphere}}=[3.0, 7.0, 9.0, 7.0]$ for spheres, and $\kappa_{\text{plane}}=[5.0, 13.5, 6.0, 3.0]$ for the boundary planes.


\subsubsection{Evaluation}
We report the success rate over the $100$ environments of each clutter level in Fig. \ref{fig:statistical_result}. The CBF constraints are constructed using $r_q$, inflated by a padding factor of $3$, where $r_q$ denotes the radius of the smallest sphere enclosing the quadrotor. This inflation defines a collision boundary that accounts for uncertainties in the model and state estimation.


We count a trial as a success when the quadrotor reaches and stops within $r_q$ of the goal position without colliding with any obstacle. 

Fig. \ref{fig:statistical_result} reports the outcome of all $300$ trials. The fourth-order CBF-QP reaches the goal in $88$ of $100$ trials at $n=12$, in $84$ of $100$ at $n=20$, and in $89$ of $100$ at $n=28$, and no trial at any clutter level ends in a collision. The success rate varies by at most five points across the three levels and does not fall monotonically with $n$, which indicates that the fourth-order constraint remains feasible as the obstacle count more than doubles.

In simulation experiments, every failure is a deadlock rather than a collision. 
In a deadlock, the quadrotor settles at an undesired equilibrium on the boundary of an obstacle and never reaches the goal, so the run violates the goal criterion while the safety constraint still holds. Deadlock is a known limitation of CBF-based safety filters \cite{deadlock}, and a detailed analysis of its causes appears in \cite{deadlock2}. We leave its resolution to future work.

\subsection{Real-World Experiments}\label{sec:real-world}
We conducted three real-world flights in the cluttered environment shown in Fig. \ref{obs_env}. The quadrotor platform is a customized quadrotor equipped with an NVIDIA Jetson Orin NX computer and a Matek H743 autopilot board. We performed the flight in a netted space in which a Vicon motion capture system provides the vehicle state at $240\,\mathrm{Hz}$. The nominal NMPC and the fourth-order CBF-QP safety filter run as ROS2 C++ nodes on the onboard computer at $250\,\mathrm{Hz}$, while the autopilot executes the thrust and torque command at $2000\,\mathrm{Hz}$.


The reference is a straight-line trajectory at a fixed height of $1.5\,\mathrm{m}$, from $(0,0,1.5)$ to $(5.5,0.2,1.5)$ in the world frame. The environment contains nine cylindrical obstacles that represent trees and barrels. Their horizontal centers and radii, in meters, are $(2.2,0.0;0.4)$, $(2.5,-1.6;0.4)$, $(2.3,1.6;0.4)$, $(3.2,0.4;0.4)$, $(3.65,0.0;0.4)$, $(4.3,0.1;0.4)$, $(4.8,-1.7;0.3)$, $(5.3,-1.4;0.3)$, and $(4.2,-1.5;0.3)$, where each tuple is $(c_x,c_y;r)$. Each obstacle is inflated by the quadrotor safety margin before constructing the CBF constraints. The resulting experimental layout is illustrated in Fig. \ref{obs_env}. 

Fig. \ref{fig:3d} shows the results of the real-world experiments. The quadrotor tracks the straight-line reference until an obstacle constraint becomes active, and the filter laterally deflects it around the obstacle. The quadrotor subsequently continues toward the goal region after clearing the obstacle. All three reported flights reach the goal region without physical collision, and the CBFs remain nonnegative throughout the reported trajectories, even though the closed-loop system ended with different trajectories in different flights. Figure \ref{fig:cbf_cons} shows all the CBF values corresponding to the obstacles and the bounding boxes, where the minimum CBF values are marked in blue. 
The minimum CBF values across the three flights shown in Fig. \ref{fig:cbf_cons} are $0.00311$, $0.02958$, and $0.05559$, respectively, which confirms the collision-free flights. The actual distance corresponding to the minimum CBF value $0.00311$ is around $0.055$ m, which corresponds to one of the tree obstacles. Due to the conservative geometrical modeling of the tree, there is no physical collision between propellers and leaves in this tight scenario.

\section{Conclusions}
In this work, we propose a fourth-order CBF considering the full quadrotor dynamics for the quadrotor obstacle avoidance tasks, and we apply the integral HOCBF method to resolve the mixed-relative degree issue. We incorporate the derived high-order CBF condition in a CBF-QP as a safety filter along with the NMPC nominal controller in a cascaded structure. We tested the control design in simulation and real-world experiments to verify the effectiveness of the proposed methods. The simulation results demonstrate collision-free operation across the tested environments, and all three real-world flights were completed without physical collision. A future direction is to integrate our proposed methods with onboard sensing and test the methods in more challenging outdoor cluttered environments with onboard sensing and computation only.



\bibliographystyle{IEEEtran}
\bibliography{ref}
\end{document}